\pdfoutput=1

\documentclass[11pt]{article}

\usepackage[final]{acl}

\usepackage{times}
\usepackage{float}
\usepackage{latexsym}
\usepackage{makecell}
\usepackage{amssymb}
\usepackage{stfloats}
\usepackage{booktabs}
\usepackage{caption}
\usepackage[bottom]{footmisc}

\usepackage[T1]{fontenc}

\usepackage[utf8]{inputenc}

\usepackage{microtype}

\usepackage{inconsolata}

\usepackage{graphicx}
\usepackage{multirow}

\newcommand{\tighttt}[2][10pt]{%
  {\fontsize{#1}{#1}\selectfont
   \ttfamily
   \spaceskip=0.3em\relax
   #2%
  }%
}

\title{Reasoning-Enhanced Domain-Adaptive Pretraining of Multimodal Large Language Models for Short Video Content Governance}

\author{
    Zixuan Wang\thanks{Equal contribution.}, Yu Sun\footnotemark[1], Hongwei Wang, Baoyu Jing, Xiang Shen, Xin Dong,\\\textbf{Zhuolin Hao, Hongyu Xiong, Yang Song}\\
    TikTok Inc.\\
    \{zixuan.wang1, yu.sun, hongwei.w, baoyu.jing, xindong, \\haozhuolin, hongyu.xiong\}@tiktok.com
}

\begin{document}
\maketitle
\begin{abstract}
Short video platforms are evolving rapidly, making the identification of inappropriate content
increasingly critical.
Existing approaches typically train separate and small classification models for each type of issue, which requires extensive human-labeled data and lacks cross-issue generalization.
We propose a reasoning-enhanced multimodal large language model (MLLM) pretraining paradigm for unified inappropriate content detection. To address the distribution gap between short video content and the original pretraining data of MLLMs, as well as the complex issue definitions, we introduce three targeted pretraining tasks:
(1) \textit{Caption}, to enhance the MLLM's perception of video details;
(2) \textit{Visual Question Answering (VQA)}, to deepen the MLLM's understanding of issue definitions and annotation guidelines;
(3) \textit{Chain-of-Thought (CoT)}, to enhance the MLLM's reasoning capability.
Experimental results show that our pretraining approach significantly improves the MLLM's performance in both zero-shot and supervised fine-tuning (SFT) settings.
In addition, our pretrained model demonstrates strong generalization capabilities to emergent, previously unseen issues.

\end{abstract}

\section{Introduction}

\begin{figure*}[h] 
  \centering
  \includegraphics[width=1\textwidth]{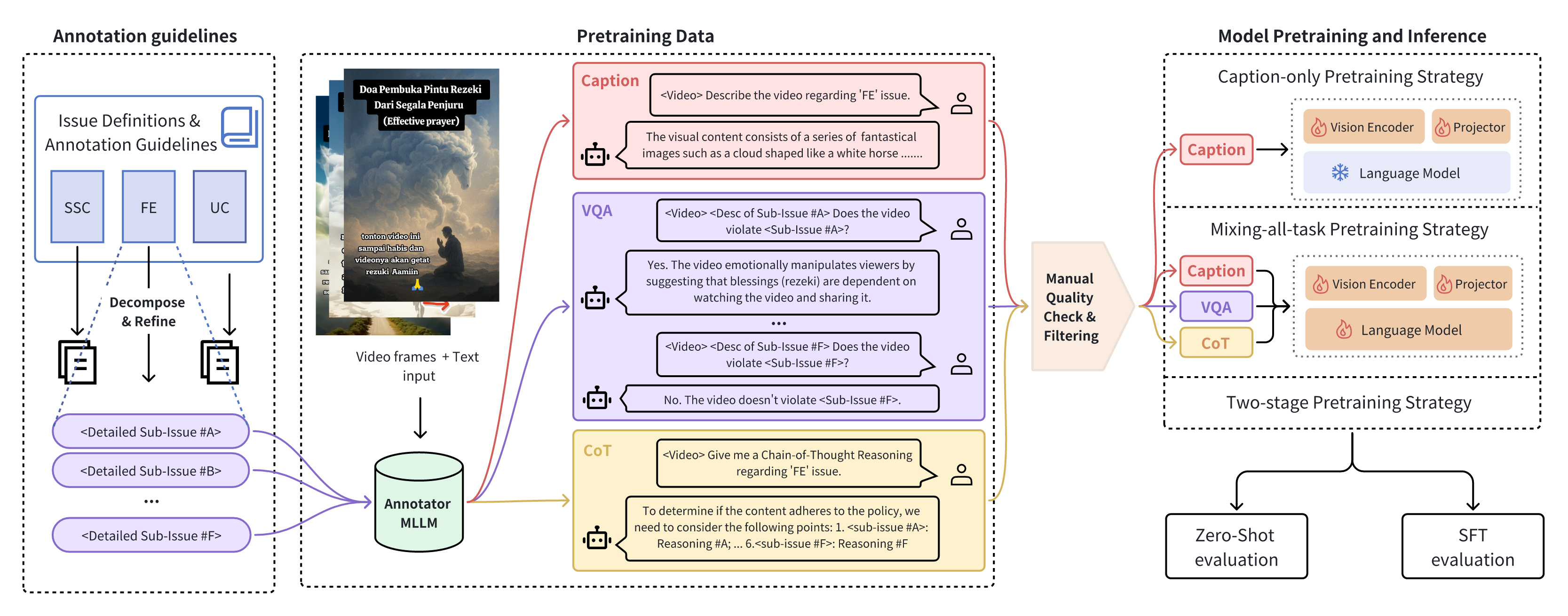}
  \caption{Illustration of our domain-adaptive pretraining approach for short video content governance. For each issue type, we first decompose the annotation guidelines into a set of sub-questions to assist pretraining data generation. An annotator MLLM is deployed to produce three types of pretraining data: Caption, VQA, and CoT, enabling the model to mimic human-like reasoning. The model can be pretrained using three different strategies, and the pretrained model is finally evaluated in both zero-shot and SFT settings.}
  \label{fig:flow}
\end{figure*}

Short video platforms such as Reels and YouTube Shorts have experienced explosive growth.
While these platforms offer unprecedented opportunities for communication and information dissemination, they have simultaneously facilitated the distribution of a wide range of inappropriate content.
Videos featuring borderline sexual content, 
plagiarism, and fake engagement, not only 
affect user
experience but also deteriorates content ecosystem in the long run.
As a result, the regulation and governance of short video content have become increasingly critical to ensure a healthy digital ecosystem.

Current short video platforms typically implement a standardized content governance process, which consists of three sequential stages:
The platform initially develops comprehensive \textit{guidelines} for identifying specific \textit{issues} (such as sexually suggestive content);
Subsequently, human annotators are instructed to label video samples according to these guidelines;
Finally, this labeled dataset serves as labeled data to fine-tune visual models (e.g., VLMo \cite{bao2022vlmo}, BEiT \cite{bao2022beitbertpretrainingimage}, and X-VLM~\cite{wang2022multi}) for video classification, which are then deployed for real-time content monitoring and filtering.
However, this process comes with significant limitations:
(1) \textit{High manual labeling cost}.
These visual models are typically small in size (usually only a few hundred megabytes) and lack world knowledge, so they require large volumes of manually labeled data, making the process time-consuming and expensive.
(2) \textit{No cross-issue generalization}.
Each of these small models is designed to handle only a fixed issue and lacks the ability to generalize across different issues.
Therefore, when the definition of an issue changes or a new issue emerges, the model must be retrained from scratch following the same labor-intensive process, leading to long development cycles.

To address the limitations of small specialized models, we explore the use of multi-modal large language models (MLLMs).
MLLMs possess extensive world knowledge, which theoretically reduces the need for large-scale annotated data and offers better cross-issue generalization.
However, applying MLLMs in this context presents two key challenges:
(1) \textit{Domain-specific data distribution}.
The data on short video platforms exhibits unique characteristics, both in \textit{format} (e.g., frame layouts, visual composition, and hashtag usage patterns) and in \textit{semantics} (e.g., narrative styles, editing conventions, and internet slang).
These characteristics often diverge significantly from the distribution of data MLLMs were originally pretrained on, potentially impacting performance.
(2) \textit{Complex and evolving definition of issues}.
Each content issue comes with its own well-defined criteria and decision rules, which are usually complicated and may evolve.
For an MLLM to make accurate judgments, it must thoroughly understand and follow these guidelines to perform reliable inference.

To tackle the above challenges, we propose a novel domain-adaptive pretraining approach for short video governance.
As illustrated in Figure \ref{fig:flow}, the pretraining approach includes three key components:
(1) \textit{Enhanced video detail perception}.
To help MLLMs better adapt to the unique data patterns of short videos, we introduce a descriptive video captioning (\textit{Caption}) task into the pretraining stage.
These captions guide the model to attend to fine-grained video detail, especially those relevant to specific issues.
(2) \textit{Deep understanding of guidelines}.
To enable MLLMs to thoroughly understand the nuances of issue annotation guidelines, we decompose each guideline into a set of sub-questions and construct a visual question answering (\textit{VQA}) task.
For example, in the case of borderline sexual content, the guideline is broken down into questions such as ``\tighttt{Are private body parts exposed?}'', with annotated answers.
This enables MLLMs to learn the specific judgment criteria for each issue.
(3) \textit{Structured reasoning}.
Even after perceiving video details and understanding the guidelines, the model must follow a systematic reasoning process.
To support this, we curate a chain-of-thought (\textit{CoT}) task that teaches MLLMs to reason step by step following predefined logic flows.
Together, the \textit{Caption}, \textit{VQA}, and \textit{CoT} datasets provide end-to-end domain adaptation for MLLMs, which significantly improves the performance.

We conducted the pretraining on several open-source MLLMs, including LLaVA-OV \cite{li2024llava} and Qwen2.5-VL \cite{qwen2.5vl2024}, using data from three content governance tasks: sexually suggestive content, unoriginal content, and fake engagement.
Experimental results show that our domain-adaptive pretraining significantly improves model performance compared with native models, both in zero-shot and supervised fine-tuning (SFT) settings.
Moreover, we demonstrate that our approach enables strong generalization to unseen issues with minimal supervision, highlighting its robustness and adaptivity.
Our ablation studies further validate the effectiveness of our method across different model scales.

The contribution of this paper is as follows:
\begin{itemize}
    \item We propose an MLLM pretraining framework for short video governance, demonstrating strong generalization to out-of-distribution issues not seen during pretraining.
    \item We design three pretraining tasks (Caption, VQA, CoT) to enhance the model’s ability to perceive video details, understand annotation guidelines, and perform structured reasoning.
    \item Experiments show that our pretrained model significantly improves performance in both zero-shot and SFT evaluation settings.
\end{itemize}

\section{Related Work}
    \subsection{Video Content Governance}
    Recent advances in deep neural networks have significantly enhanced content governance capabilities in detecting and classifying inappropriate video content on short video platforms \cite{yousaf2022deep, li2025cogvla}.
    For example, multimodal features have been applied to identify misleading clickbait videos \cite{rahman2023identification, sun2025audio, liang2025emb};
    Age-adaptive learning techniques have been developed to detect and categorize inappropriate video content for different demographics of viewers, demonstrating sensitivity to visual needs \cite{alam2024utilizing, DBLP:conf/aaai/ZhangLXCW024, li2025lion}.
    These approaches underscore the potential of MLLMs in addressing the challenges of short video content governance.
    
    \vspace{0.5em}


    \subsection{Domain-Adaptive Pretraining}
    Recent research demonstrates the effectiveness of domain-adaptive pretraining in enhancing model performance \cite{song2025injecting}.
    Specifically, pretraining strategies have been used in model design, task formulation, and application scenarios \cite{guo2024efficient}.
    For example, in multimodal learning, efficient domain-specific pretraining has been applied to human activity recognition \cite{bulat2024efficient} and short video understanding \cite{lu2025vlmpolicycommonlawcontent,li2023ultrare};
    SimRAG \cite{hong2024intelligent} and domain-specific instruction tuning \cite{xie2024efficient, liu2024agentps, wang-etal-2025-filter, 10.1145/3711896.3737203} effectively align models with VQA tasks \cite{ging2024open, khullar2024improved};
    Adaptive techniques such as velocity-based domain reweighting \cite{luo2024velocitune,leong2025amasadaptivelydeterminingcommunication}, knowledge distillation \cite{10.1145/3627673.3680045} and structure-aware knowledge injection \cite{liu2024structure} further improve knowledge retention and transfer.

\section{The Proposed Pretraining Approach}
    \subsection{A Unified Model for All Content Issues}
        Short video platforms often face a variety of content issues, such as non-original content.
        We denote the set of all issues of interest by $\mathcal I$.
        For each issue $i \in \mathcal I$, the policy team defines detailed descriptions and corresponding annotation guidelines, denoted by $G_i$.
        Human annotators are then trained using these guidelines to label a collection of videos, resulting in a dataset $\mathcal D_i$.
        This dataset is then used to fine-tune a visual classification model $\mathcal M_i$:
        \begin{equation}
            \mathcal M_i (v; G_i, \mathcal D_i) \rightarrow l, \quad \textrm{for} \ i \in \mathcal I,
        \end{equation}
        where $v$ is an input short video and $l$ is the predicted label for $v$.
        
        However, due to the lack of cross-issue generalization in these models, any change in the annotation guidelines $G_i$ or the emergence of a new issue would require collecting a new human-annotated dataset $\mathcal{D}_i$ and retraining the model from scratch, which is extremely costly.
        Therefore, a promising solution is to leverage the broad, internalized knowledge of MLLMs to pretrain a unified model $\mathcal M$ that can generalize across different issues:
        \begin{equation}
            \mathcal M(v; \{G_i\}_{i \in \mathcal I}, \mathcal D) \rightarrow l,
        \end{equation}
        where $\mathcal D$ is the pretraining dataset, which we will describe in detail in Section \ref{sec_pretraining_tasks}.

    \subsection{Annotation Guidelines Decomposition}
        \label{sec:guideline_decomposition}
        Issue definitions and annotation guidelines are usually highly complex, typically covering dozens of scenarios, their exceptions, and both positive and negative illustrative examples.
        To help the model better understand these guidelines, it is essential to simplify and distill them into a set of sub-questions that are easier for the model to process.
        For instance, the annotation guidelines for the issue of sexually suggestive content consist of intricate definitions: ``\tighttt{Adult Image-Based Sexual Abuse occurs when the subject(s) depicted ...}''.
        We decompose them into a collection of simple sub-questions such as ``\tighttt{Are private body parts exposed?}'', ``\tighttt{Is there sexual teasing or invitation?}'', and ``\tighttt{Are adult products shown?}''.
        These sub-questions are later used to construct the pretraining dataset.

    \subsection{Pretraining Tasks}
        \label{sec_pretraining_tasks}

        There are two main challenges in using MLLMs for short video content governance:
        (1) The unique characteristics of short videos limit the capabilities of native MLLMs to fully understand their content;
        (2) Issue annotation guidelines are often complex, making it difficult for native MLLMs to effectively follow and reason based on them.
        To address these challenges, we design three pretraining tasks:

        \textbf{Caption} task is to help the model perceive fine-grained details of the input video, especially those related to specific issues.
        As shown in Figure \ref{fig:flow}, the input of the caption task is a prompt such as ``\tighttt{Describe the video regarding the fake engagement issue.}'', and the expected output is a 2–3 sentence description of the video from the perspective of fake engagement.

        \textbf{VQA} task is to enable the model to develop a deep understanding of the annotation guidelines.
        To make more efficient use of the VQA data, we design two usage strategies:
        (1) \textit{Binary QA}, where the input is a sub-question derived from the decomposed annotation guidelines described in Section \ref{sec:guideline_decomposition}.
        The output is a yes/no answer of whether the input video violates the given sub-question, as well as a detailed explanation.
        Examples of binary QA task are illustrated in Figure \ref{fig:flow}.
        (2) \textit{Multi-choice QA}, where the input consists of all sub-questions, and the model is asked to select which issues the input video violates from the given options.
        It is important to note that VQA task relies on the model’s accurate perception of video details obtained through the Caption task.

        \textbf{CoT} task is to enable the model to integrate the answers from all sub-questions in the VQA task and produce a complete reasoning process leading to a final conclusion.
        As shown in Figure \ref{fig:flow}, for the fake engagement issue, the model should go through all the sub-questions.
        The final conclusion will be positive if any of the answer are positive.

    \subsection{Model Pretraining and Inference}
        How to organize the three pretraining tasks is also a critical question. A straightforward way is to mix all three tasks and jointly train the model.
        However, we found that the caption task is particularly important. Performing an initial round of caption-only training followed by joint training of all three tasks leads to better performance.
        We discuss the pretraining recipes in detail in Section \ref{sec:training_recipes}.

        We design two inference strategies.
        (1) The first is \textit{zero-shot}, where the pretrained model directly classifies the input video and provides an explanation.
        We use the normalized logits of the predicted label token as the output probability.
        (2) The second strategy is \textit{SFT}, where we add an MLP classification head on top of the model’s final layer, and fine-tune the model using LoRA \cite{hu2022lora} on an additional SFT dataset for classification.
        It’s worth noting that zero-shot inference offers more interpretable explanations for the model’s predictions, while SFT inference is faster and currently serves as the solution for online deployment.

\section{Experimental Setup}

    \subsection{Content Issues}
    We collect the pretraining data and conduct experiments on real industrial data spanning three issues: Sexually Suggestive Content (\textbf{SSC}), Unoriginal Content (\textbf{UC}), and Fake Engagement (\textbf{FE}).
    To evaluate the model’s ability to generalize to unseen or out-of-distribution issues, we also test the pretrained model on the Shocking Graphic Content (\textbf{SGC}) issue, which includes videos that cause physical discomfort to viewers.
    SGC issue is not included in the pretraining data.

\begin{table*}[t]
\centering
\scriptsize
\setlength{\tabcolsep}{8pt}

\begin{tabular}{c|c|cc|cc|cc|cc|c}
\toprule
\multirow{2}{*}{\textbf{Model}} & \textbf{Pretraining} & \multicolumn{2}{c|}{\textbf{SSC}} & \multicolumn{2}{c|}{\textbf{UC}} & \multicolumn{2}{c|}{\textbf{FE}} & \multicolumn{2}{c|}{\textbf{SGC}} & Overall \\

& \textbf{strategy} & ACC & F1 & ACC & F1 & ACC & F1 & ACC & F1 & AUC \\
\midrule
\midrule
LLaVA-OV native & - & 70.68 & 70.62 & 49.60 & 66.31 & 48.66 & 65.20 & 71.19 & 76.20 & 61.97 \\
\midrule
\multirow{3}{*}{LLaVA-OV pretrained} 
& Caption & 74.53 & 75.38 & 65.67 & 71.82 & 48.09 & 64.95 & 71.19 & 75.31 & 70.89 \\
& Mix     & \textbf{82.13} & \underline{82.34} & \underline{78.87} & \underline{79.10} & \underline{75.81} & \textbf{77.39} & \underline{75.13} & \underline{77.49} & \underline{80.75} \\
& Stage   & 80.65 & 81.99 & \textbf{80.46} & \textbf{80.74} & 71.13 & 72.69 & \textbf{77.20} & \textbf{78.76} & \textbf{81.22} \\
\midrule
Qwen2.5-VL native & - & 74.23 & 76.12 & 51.69 & 67.12 & 53.82 & 66.44 & 74.92 & 78.04 & 69.49 \\
\midrule
\multirow{3}{*}{Qwen2.5-VL pretrained} 
& Caption & 76.51 & 76.53 & 51.59 & 66.53 & 49.33 & 65.22 & 73.89 & 76.14 & 68.82 \\
& Mix     & \underline{81.64} & \textbf{82.49} & 75.40 & 75.59 & 73.33 & 74.33 & 72.12 & 76.17 & 74.10 \\
& Stage   & 80.75 & 81.55 & 76.09 & 76.40 & \textbf{76.00} & \underline{75.12} & 73.89 & 76.00 & 74.24 \\
\midrule
GPT-4o & - & 75.71 & 72.97 & 65.97 & 67.24 & 64.15 & 68.97 & 74.62 & 78.09 & - \\
\bottomrule
\end{tabular}

\caption{Result of zero-shot evaluation. Metrics are accuracy and F1 (in \%). The last column is the overall ROC-AUC (in \%) across all issues. The highest value in each column is shown in \textbf{bold}, and the second-highest is \underline{underlined}. The overall AUC of GPT-4o cannot be computed because we cannot access its output probabilities.}
\label{tab:zs}
\end{table*}

\subsection{Datasets}

    \textbf{Pretraining data}.
    We 
sample 50k positive and 50k negative videos for each of the 3 issues.
    An MLLM annotator is used to generate the Caption, VQA, and CoT tasks.
    The Caption task is generated independently, while the VQA and CoT tasks are generated simultaneously, resulting in a total of 920k instruction samples.
    After generation, we manually filter out any VQA-CoT samples that are inconsistent with the human annotations.


    \noindent \textbf{SFT data}.
    We use 
    human-annotated binary-labeled examples as SFT data for each issue to fine-tune the pretrained model, with 10\% of the samples labeled as positive.
    
    \noindent \textbf{Evaluation data}.
    We use 1k human-annotated samples as evaluation data for each issue, with 50\% of the samples labeled as positive.


\subsection{Baseline Models}
    We choose LLaVA-OV 7B \cite{li2024llava} and Qwen 2.5-VL 7B\cite{qwen2.5vl2024} as baseline MLLMs for continual pretraining.
    Both models consist of three components: a language model, a vision encoder, and a projector.
    Additionally, we compare our pretrained models with GPT-4o \cite{achiam2023gpt} in the zero-shot setting.


\subsection{Pretraining Strategy}
    \label{sec:training_recipes}
    We use three pretraining strategies:
    
    \noindent \textbf{Caption-only}.
    In this strategy, the language model is frozen while the vision encoder and projector are trained using only the caption dataset.
    The goal is to enhance the model’s ability to perceive fine-grained visual details.

    \noindent \textbf{Mixing-all-tasks}.
    All three types of pretraining datasets are mixed and randomly shuffled.
    We train all components of the MLLM jointly on this mixed dataset.

    \noindent \textbf{Two-stage}.
    This is a two-stage approach that combines the previous two strategies.
    We first apply Caption-only strategy to train the vision encoder and projector, and select the best checkpoint across epochs.
    Then, we switch to the Mixing-all-tasks strategy to fine-tune the entire model.

\begin{table*}[ht]
\centering
\scriptsize
\setlength{\tabcolsep}{6pt}

\begin{tabular}{c|c|c|c|c|c|c|c|c|c|c|c|c}
\toprule
\multicolumn{1}{c|}{\multirow{2}{*}{\textbf{Model}}} & \multicolumn{3}{c|}{\textbf{SSC}} & \multicolumn{3}{c|}{\textbf{UC}} & \multicolumn{3}{c|}{\textbf{FE}} & \multicolumn{3}{c}{\textbf{SGC}}\\ 
& AUC & ACC & P@R90 & AUC & ACC & P@R90 & AUC & ACC & P@R90 & AUC & ACC & P@R90  \\ 
\midrule
\midrule
Native & 92.97 & 84.05 & 79.76 & 80.98 & 71.53 & 59.92 & 91.38 & 84.35 & 75.70 & 99.18  & 89.86 & 97.95 \\
\midrule
Pretrained w/ Mix   & \textbf{93.29} & \textbf{85.52} & \textbf{81.57} & \textbf{87.53} & \textbf{75.79} & \textbf{69.88} & \textbf{91.94} & \textbf{84.46} & \textbf{78.31} & 99.26 & 94.93 & \textbf{98.17}  \\
\midrule
Pretrained w/ Stage   & 93.00 & 84.84 & 80.57 & 86.65 & 73.41 & 64.84 & 91.88 & 84.34 & 76.12 & \textbf{99.31} & \textbf{95.14} & 98.05  \\ 
\bottomrule
\end{tabular}

\caption{Result of SFT evaluation. Metrics are ROC-AUC, Accuracy, and P@R90 (Precision at 90\% Recall) (in \%). Following the actual model deployment setup, LLaVA-OV is used as the base model for SSC, FE, and SGC, while Qwen2.5-VL is used as the base model for UC. The highest value in each column is shown in \textbf{bold}.}
\label{tab:sft_llava}
\end{table*}


\section{Experimental Results}

    \subsection{Zero-Shot Evaluation}
        \textbf{In-domain issues}.
        The results of zero-shot evaluation are shown in Table \ref{tab:zs}.
        For the three in-domain issues: SSC, UC, and FE, the performance of the pretrained models consistently surpasses that of their native counterparts.
        Notably, the pretrained models also significantly outperform GPT-4o.
        This indicates that, despite GPT-4o's strong multimodal capabilities, it still struggles to fully understand complex issue annotation guidelines.

        From an issue-specific perspective, the accuracy improvements brought by pretraining on SSC, UC, and FE are approximately 7–11\%, 24\%–31\%, and 22\%, respectively.
        The gains on UC and FE are greatly higher than those on SSC.
        This may be because SSC is detecting sexually suggestive content, a task for which native models already possess sufficient prior knowledge.
        In contrast, UC and FE are non-original content and fake engagement, whose precise definitions are given in the annotation guidelines.
        Native models lack strong priors for these issues.
        This further demonstrates the effectiveness of our VQA and CoT pretraining tasks in injecting domain-specific knowledge into the base models.

        \vspace{0.5em}
        
        \noindent \textbf{Out-of-domain issue}.
        To demonstrate the generalization ability of the pretrained models, we also evaluate their performance on the SGC issue, which is not included in the pretraining data.
        The results show that pretraining significantly improves LLaVA’s performance on this out-of-domain issue ($6\%$ absolute accuracy improvement).

        \vspace{0.5em}
        
        \noindent \textbf{Comparison between LLaVA and Qwen}.
        On the three in-domain issues, pretraining brings accuracy improvements of 11\%–31\% for LLaVA and 7\%–24\% for Qwen.
        This suggests that the benefits of pretraining are more significant for LLaVA.
        Upon closer inspection, we found that one possible reason is that Qwen sometimes produces inconsistencies between its reasoning and final answers.
        That is, it may generate a correct and well-structured reasoning process, but still output an incorrect final prediction.
        We provide a detailed analysis example in the Appendix.

        \vspace{0.5em}
        
        \noindent \textbf{Pretraining strategy}.
        Comparing the three pretraining strategies, we find that Two-stage consistently outperforms Caption-only and Mixing-all-tasks in most cases.
        This suggests that first training the model to perceive fine-grained video details, followed by learning the annotation guidelines and reasoning process, is the most effective practice.

\begin{table}[t]
\centering
\scriptsize
\setlength{\tabcolsep}{5pt}
\renewcommand{\arraystretch}{1.05}

\begin{tabular}{c|c|c|c}
\toprule
\textbf{Configuration} & \textbf{AUC} & \textbf{ACC} & \textbf{P@R90} \\ 
\midrule
\midrule
LLaVA-OV native + 100\% SFT data & 99.18  & 89.86 & 97.95  \\
\midrule
LLaVA-OV pretrained + 100\% SFT data & \textbf{99.31} & 95.14 & 98.05 \\
\midrule
LLaVA-OV pretrained + 50\% SFT data & 99.26 & \textbf{95.35} & \textbf{98.17} \\
\bottomrule
\end{tabular}
\caption{ROC-AUC, Accuracy, and P@R90 scores (in \%) on the SGC issue under SFT evaluation. The pretraining strategy is Stage. The highest value in each column is shown in \textbf{bold}.}
\label{tab:data_size}
\end{table}

    \subsection{SFT Evaluation}
        The SFT evaluation results are presented in Table \ref{tab:sft_llava}.
        Compared to zero-shot evaluation, most of our conclusions remain consistent, with a few differences:

        (1) In most cases, pretraining still improves model performance under SFT evaluation, but the gains are generally smaller, typically around 1\% to 10\%.
        This is because the pretraining paradigm aligns closely with the zero-shot evaluation format (i.e., natural language generation), whereas the SFT evaluation is direct classification through probability outputs.
        This mismatch reduces the impact of pretraining under SFT evaluation.

        (2) Regarding pretraining strategies, Mixing-all-tasks performs slightly better than Two-Stage.
        Note that Two-Stage includes an additional round of caption-only training.
        This suggests that adding an extra caption phase does not significantly enhance performance under SFT evaluation.
        The reason is similar to the above: the Caption task aligns more closely with natural language generation and contributes less to performance when the evaluation requires direct classification.
    



\subsection{Ablation Study}

\textbf{Cross-issue generalization}.
Pretrained MLLMs exhibit strong cross-issue generalization capabilities.
This generalization is reflected not only in the zero-shot setting, where pretrained models perform well on out-of-domain issues (as shown in the SGC issue in Table 1), but also in the SFT setting, where pretrained models can achieve or even surpass the performance of native models using significantly less SFT data.
To demonstrate this, we fine-tune the pretrained LLaVA model using only 50\% of the SFT data.
The results, shown in Table \ref{tab:data_size}, reveal that the pretrained LLaVA with just half the SFT data outperforms the native LLaVA trained with the full SFT dataset (compare row 3 with row 1).
This highlights that the pretrained model possesses substantially stronger cross-issue generalization capabilities than the native model, enabling it to adapt to new issues with minimal SFT data.

\vspace{0.5em}

\begin{figure}[t]
    \centering
    \includegraphics[width=0.8\linewidth]{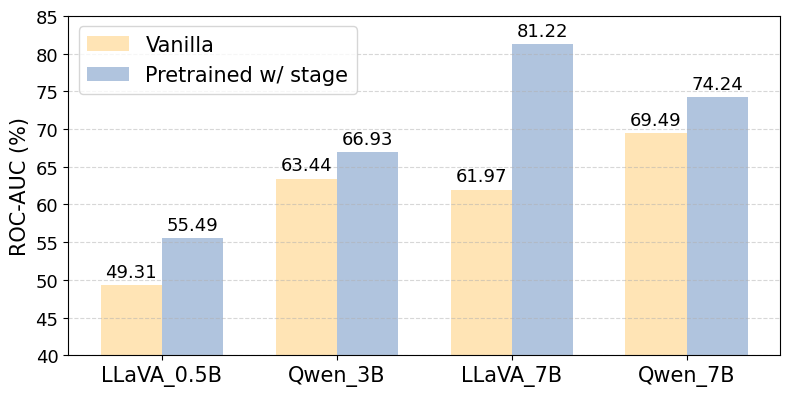}
    \caption{ROC-AUC of vanilla and pretrained models with different model sizes under zero-shot evaluation.}
    \label{fig:model_size}
\end{figure}

\noindent \textbf{Base model size}.
    To investigate the impact of base model size on pretraining effectiveness, we additionally conduct pretraining and zero-shot evaluation on 0.5B versions of LLaVA and Qwen.
    The results in Figure \ref{fig:model_size} demonstrate that our pretraining approach also yields significant performance improvements (3\%–6\%) for 0.5B models.
    Moreover, we observe that the performance gains on 7B models (5\%–19\%) are higher than those on 0.5B models.
    This may be because larger models have greater capacity and can benefit more from the same amount of pretraining data, allowing them to reach a higher performance ceiling.
 
\begin{figure*}[t]
    \centering
    \includegraphics[width=\linewidth]{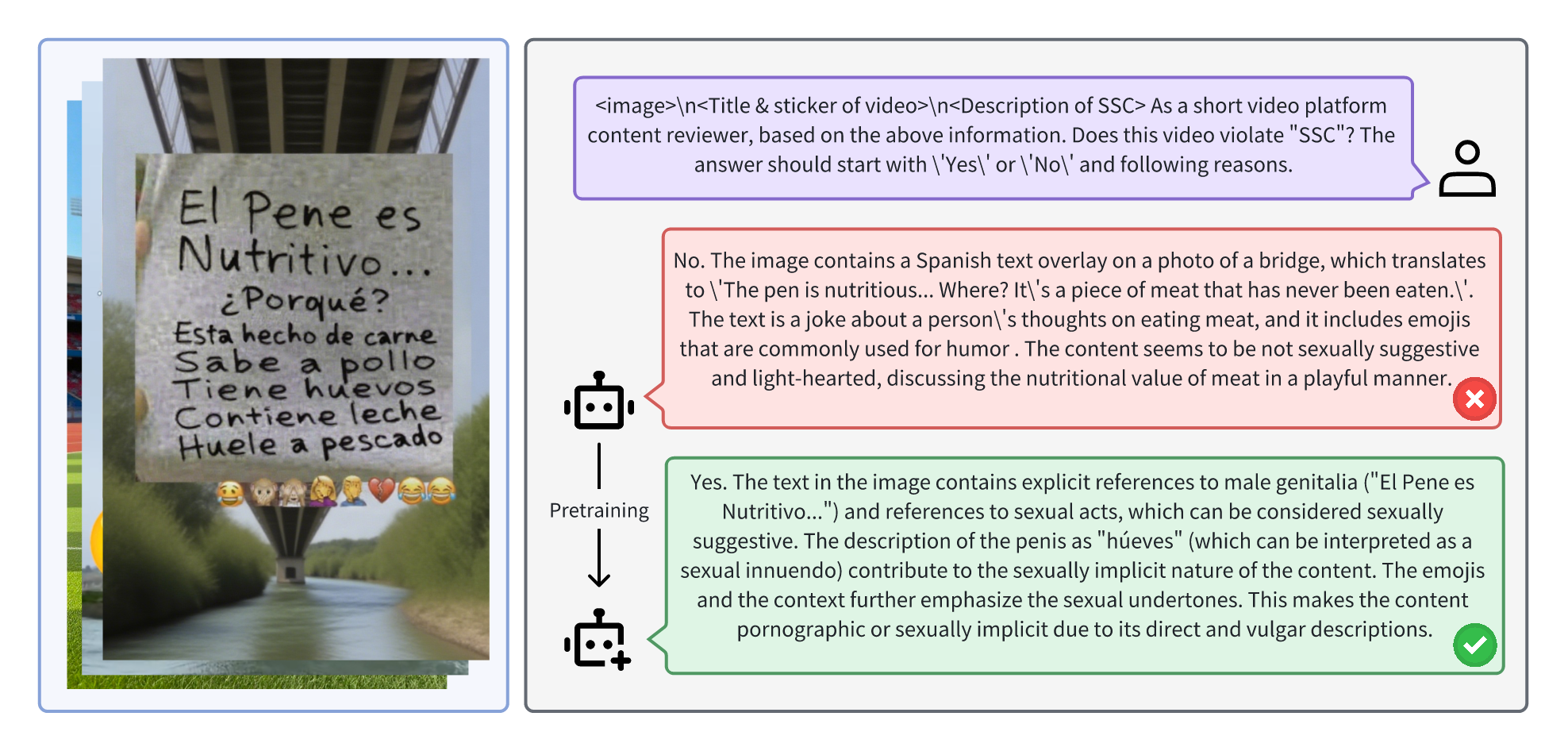}
    \caption{Left: An illustrative example of short video frames that violates the SSC policy. Right: Zero-shot evaluation prompt and the MLLM’s responses before and after our pretraining.}
    \label{fig:case_study}
\end{figure*}

\subsection{Case Study}
    We conducted a case study to intuitively demonstrate the effect of pretraining.
    As shown in Figure \ref{fig:case_study}, the Spanish text in the video frame contains an obvious sexually suggestive message.
    However, the model without pretraining fails to recognize the meaning of the text and subsequently produces incorrect reasoning.
    In contrast, the pretrained model provides both an accurate answer and a correct reasoning process.


\section{Conclusion}
    In this work, we propose a domain-adaptive pretraining approach for short video content governance.
    We design three novel pretraining tasks: Caption, VQA, and CoT, which effectively enhance MLLMs' ability to perceive video details, understand annotation guidelines, and perform reasoning.
    Experimental results show that our pretraining method significantly improves the model's performance on both zero-shot and SFT evaluations.
    Furthermore, the pretrained models exhibit strong generalization to new issues, which can substantially reduce manual annotation costs and shorten development cycles in real-world deployment scenarios.

\section{Limitation}
    In this work, the largest model we used is 7B due to limitations in resources.
    Although the pretrained 7B models already perform well for short video content governance, we are still interested in exploring the impact of pretraining on even larger models.
    Regarding data types used for pretraining, in addition to MLLM-annotated data, we also plan to incorporate a large amount of unlabeled data to further enhance performance.
    Moreover, when decomposing annotation guidelines into sub-questions, we currently rely on manual design, which may not be optimal.
    In the future, we plan to explore reinforcement learning methods to automatically decompose annotation guidelines in a more efficient and adaptive manner.

\bibliography{reference}

\appendix

\section{Details of Prompt Design}
    \label{sec:prompt}

\begin{figure*}[t] 
      \centering
      \includegraphics[width=1\textwidth]{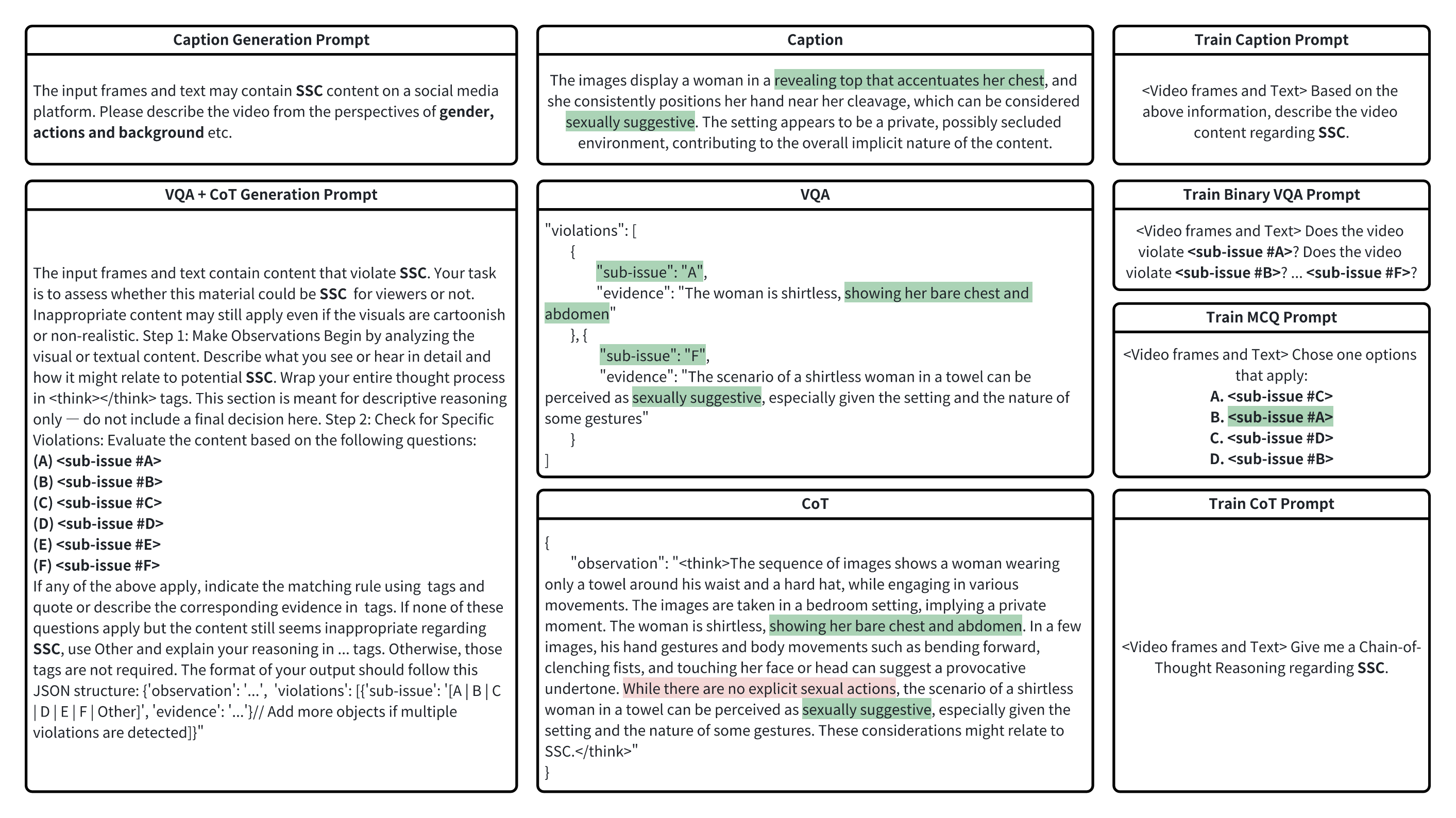} 
      \caption{An illustrative example of prompts and the generated pretraining data. The first column is the prompt for generating the pretraining data. The second column is the output generated by the prompt in the first column. The third column is the prompt used during pretraining.}
      \label{fig:prompt}
    \end{figure*}

    In Figure \ref{fig:prompt}, we present a complete pipeline example for constructing pretraining data, including the prompts used for generating pretraining data, the generated outputs, and the prompts used during pretraining. To improve clarity, we highlight the issue and its related content using bold text. Additionally, we mark segments of the generated pretraining data that align with the sub-issue in green, and those that do not in red. This clearly illustrates that issue-specific knowledge is consistently embedded across all tasks.

\section{Analysis on Qwen's Inconsistency Issue}
During zero-shot inference, we observed that when the answer format is designed as ``Reason after Answer'', Qwen often produces incorrect final answers despite providing correct reasoning.
For example, when given a video that does not violate the UC policy and asked whether it violates the UC policy, the model responds:
``Yes. The images appear to be original user-generated content without any visible watermarks indicating ownership by another entity...''
Although the model clearly identifies the content as original, its final answer is still ``Yes'', and the probability score for the first token exceeds 0.67.
As a result, evaluating performance based solely on the first token’s probability leads to an underestimation of the model’s true reasoning ability.

\end{document}